\documentclass{article}

\PassOptionsToPackage{numbers, compress}{natbib}

\usepackage[preprint]{neurips_2025}

\usepackage[utf8]{inputenc} 
\usepackage[T1]{fontenc}    
\usepackage{amsmath,amssymb}
\usepackage{url}            
\usepackage{booktabs}       
\usepackage{amsfonts}       
\usepackage{nicefrac}       
\usepackage{subfiles}
\usepackage{microtype}      
\usepackage{wrapfig}
\usepackage{makecell}
\usepackage{pifont}
\usepackage{arydshln}
\usepackage{placeins}

\usepackage{amsthm}
\usepackage{enumitem}
\usepackage{comment}
\usepackage[font=small,labelfont=bf]{caption}
\usepackage{subcaption}
\newtheorem{remark}{Remark}[section]
\newcolumntype{G}{>{\color{gray}}c}

\newcommand{\EE}{\mathbb{E}}
\appendix
\setlist[itemize]{leftmargin=0pt, labelsep=1em, itemsep=0pt, parsep=0pt}
\newtheorem{assumption}{Assumption}[section]
\newtheorem{proposition}{Proposition}[section]
\usepackage{adjustbox}  
\usepackage{graphicx}
\usepackage{multirow}
\usepackage[table]{xcolor}  
\usepackage{hyperref}
\usepackage{array}
\usepackage{siunitx}
\title{CAST: Contrastive Adaptation and Distillation for Semi-Supervised Instance Segmentation} 

\author{%
  Pardis Taghavi, Tian Liu, Renjie Li, Reza Langari, Zhengzhong Tu\thanks{Corresponding author}\\
  Texas A\&M University\\
  \texttt{\{ptgh,ltmask,renjie,rlangari,tzz\}@tamu.edu} \\
}

\begin{document}

\maketitle
\begin{abstract}
Instance segmentation demands costly per-pixel annotations and computationally expensive models. We introduce CAST, a semi-supervised knowledge distillation (SSKD) framework that compresses pre-trained vision foundation models (VFM) into compact experts using limited labeled and abundant unlabeled data. CAST unfolds in three stages: (1) domain adaptation of the VFM(s) via self-training with contrastive calibration, (2) knowledge transfer through a unified multi-objective loss, and (3) student refinement to mitigate residual pseudo-label bias. Central to CAST is an \emph{instance-aware pixel-wise contrastive loss} that fuses mask and class scores to extract informative negatives and enforce clear inter-instance margins. By maintaining this contrastive signal across both adaptation and distillation, we align teacher and student embeddings and fully leverage unlabeled images. On Cityscapes and ADE20K, our $\approx 11\times$ smaller student improves over its zero-shot VFM teacher(s) by +8.5 and +7.1 AP, surpasses adapted teacher(s) by +3.4 and +1.5 AP, and further outperforms state-of-the-art SSKD methods on both benchmarks.
\end{abstract}

\section{Introduction}
\label{sec:intro}
Pixel-level instance segmentation is notoriously expensive: annotating detailed masks can take hours per image, and training state-of-the-art detectors often requires hundreds of GPU hours, putting many applications out of reach~\cite{Cordts2016Cityscapes, he2017mask}. Recent advancements in vision foundation models (VFMs)~\cite{oquab2023dinov2, liu2024grounding, yuan2025sa2va, kirillov2023segany} have substantially expanded the capabilities of computer vision systems, achieving strong performance across diverse perception benchmarks~\cite{awais2025foundation}. 

\noindent\textbf{Motivation.} Despite remarkable achievements, foundation models still cannot serve specific downstream tasks sufficiently well due to two major issues: (1) the heavy computational overhead during deployment making these models impractical for environments with limited resources~\cite{xu2024survey}; and (2) their inherently generic nature, which leads to suboptimal performance on tasks that demand domain specific expertise~\cite{sony2025foundation}. The latter stems from foundation models being optimized to perform well across a wide variety of tasks, rather than being finely tuned for the nuanced requirements of specialized applications~\cite{bommasani2021opportunities}. This challenge is  prominent in applications that involve outdoor environments, such as autonomous driving, and indoor settings, such as robotic perception~\cite{firoozi2023foundation}. Semi-supervised knowledge distillation (SSKD) for instance segmentation seeks to compress large models into efficient student models by leveraging both limited labeled data and abundant unlabeled images. Current distillation methods either treat VFMs as fixed feature extractors with simple pseudo-labeling or focus on coarse semantic tasks, failing to exploit the rich structure of unlabeled datasets to refine per-pixel predictions. Consequently, adjacent instances remain poorly separated and accuracy degrades sharply under scarce labels.  We address these issues by adapting VFMs via self‐training to enhance pseudo-label fidelity, and by injecting an instance-aware pixel‐wise contrastive loss that leverages unlabeled data to enforce clear inter‐instance margins, yielding sharper masks and superior performance in the low‐label regime.  

{\bf Status quo.}
Knowledge distillation has evolved from task‐agnostic compression~\cite{hinton2015distilling,chen2020big} to adapting VFMs for downstream tasks.  For classification and semantic segmentation, Vemulapalli et al.~\cite{Vemulapalli2024KD} distill a VFM matching its output on an unlabeled transfer set, and SAM-CLIP~\cite{wang2024sam} fuses CLIP and SAM.  However, neither method targets per-pixel instance masks nor exploits dense self‐supervision from the unlabeled pool.  Pure semi‐supervised instance segmentation methods, such as~\cite{hu2023pseudo, berrada2024guided} train teachers from scratch, doubling GPU cost, and still produce noisy masks under scarce labels. To our knowledge, no prior work unifies VFM adaptation, unlabeled data‐driven pixel‐wise refinement, and extreme student compression for instance segmentation.

\textbf{Contributions.}
We summarize our main contributions as follows:
\begin{itemize}[nosep,leftmargin=6mm]
  \item We introduce an \emph{instance-aware pixel-wise contrastive loss} that fuses mask and class predictions to drive stronger inter-instance separation, and show how to sample negatives efficiently in an instance centric setting.
  \item We propose CAST, a SSKD pipeline with three phases: (i) adapting the foundation teacher via self-training with contrastive calibration, (ii) distilling into a compact student using a unified objective that combines supervised, pseudo-label, and pixel-wise contrastive losses, and (iii) supervised fine-tuning to reduce residual bias, unifying supervised, semi-supervised, and self-supervised signals.
  \item We conduct extensive experiments on Cityscapes and ADE20K, demonstrating that our $\approx 11\times$ smaller student improves over its zero-shot VFM teacher(s) by +8.5 and +7.1 AP, surpasses adapted teacher(s) by +3.4 and +1.5 AP, and further outperforms state-of-the-art semi-supervised instance segmentation methods under the same data splits, with lower training cost.
\end{itemize}

\section{Related Work}
\label{sec:Related Work}
\noindent\textbf{Vision Foundation Models.}
VFMs~\cite{oquab2023dinov2, liu2024grounding, ravi2024sam, yang2024depth, bochkovskii2024depth}
 have revolutionized computer vision through large scale pre-training.  In parallel, recent trends focus on combining VFMs to extend their capabilities~\cite{ren2024grounded,yuan2025sa2va}. While these models excel in open-set recognition and transfer learning, their computational demands yet hinder edge deployment. Recent efforts merge VFMs via distillation: Wang et al.~\cite{wang2024sam} unify SAM and CLIP via multi-task learning, while Zhang et al.~\cite{zhang2025accessing} distill CLIP and DINOv2 into a compact model with data distillation. We extend these paradigms by leveraging VFMs for instance segmentation, focusing on balancing robustness with computational efficiency.

\noindent\textbf{Knowledge Distillation in Vision.}
Knowledge distillation (KD) has become a ubiquitous technique to transfer knowledge from teachers with high capacity to lightweight students for efficient deployment. Early methods distilled softened logits or intermediate features~\cite{hinton2015distilling} in a task-agnostic way, while later feature-based approaches capture structured spatial cues (e.g., pixel-wise similarity, channel distributions)~\cite{rajasegaran2020self, shu2021channel}. Modern methods tackle VFMs' scale and complexity:~\cite{sun2023dime, yang2024clip} distills VFMs to impart zero-shot and multimodal capabilities, further multi-teacher approaches~\cite{jiang2024mtkd, yang2025multi} combine complementary expertise.
Vemulapalli et al.~\cite{Vemulapalli2024KD} adapt a VFM to the target task and then distill on a large unlabeled set for classification and semantic segmentation. Building on these advances in vision knowledge distillation, we posit that a strong teacher (or ensemble of teachers) can effectively guide a lightweight instance segmentation model to high performance. Our approach explicitly integrates semi-supervised learning and pixel-level contrastive signals for instance segmentation, to focus on bridging the gap between rich representation of VFMs and compact, efficient student networks.

\noindent\textbf{Semi-Supervised Learning.}
Self‐training (or pseudo‐labeling) has become a foundational paradigm in semi‐supervised learning (SSL), where a model leverages its own predictions with high confidence and iteratively refines itself~\cite{xie2020self}. This approach has proven effective across vision tasks, improving image classification performance~\cite{xie2020self} and boosting object detection accuracy when annotation budgets are tight~\cite{liu2021unbiased}. To counteract error accumulation from noisy pseudo‐labels~\cite{tarvainen2017mean} use exponential moving average of label predictions, or~\cite{cascante2021curriculum} employ curriculum labeling schemes that gradually incorporate harder examples. More recent work applies pseudo-labeling for large pre-trained models through targeted finetuning and adaptive pseudo selection strategies~\cite{gan2024erasing}. While many SSL methods focus on classification or detection, several have extended this method to dense prediction tasks~\cite{chen2021semi, yang2023revisiting}.

We study self‐training with self‐supervised contrastive learning and task‐specific adaptation. Global contrastive frameworks such as SimCLR~\cite{chen2020simple}, MoCo~\cite{chen2021empirical}, and their detection extensions~\cite{xie2021detco} established the value of large-scale visual discrimination learning. Further per-pixel contrastive approaches~\cite{wang2021dense, xie2021propagate, zhong2021pixel, wang2022contrastmask, alonso2021semi} have shown promise in retaining spatial sensitivity though they yet conflate pixels from different instances of the same class. We extend these advances by synergizing self-training and self-supervised contrastive learning, and introduce a novel instance-aware negative sampling strategy designed specifically for the demands of instance segmentation.

\begin{figure}[t]
\begin{center}
   \includegraphics[width=1\linewidth]{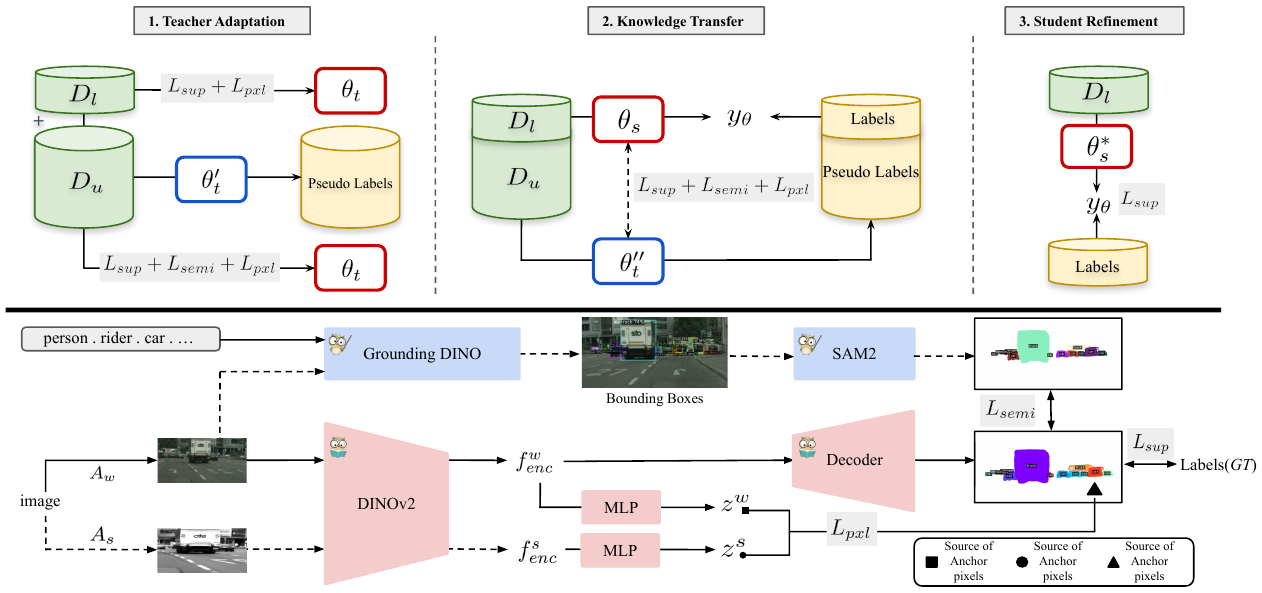}
   \end{center}
\caption{%
  \textbf{CAST framework overview.} 
  \textbf{Top:} Three-stage pipeline: 
  (1) adapt a pre-trained VFM teacher to the target domain via self-training with pixel-level contrastive calibration; 
  (2) distill knowledge into a compact student using instance-aware contrastive sampling; 
  (3) fine-tune the student on labeled data to correct residual pseudo-label bias.  
  \textbf{Bottom:} Detailed view of stage (2): fused mask and class score maps produce anchor pixels, sampled across weak/strong views to form positive/negative pairs; an MLP projects features for the contrastive loss. Dashed arrows denote no gradient flow; red modules are trainable, blue are frozen.
}

\label{fig:setup}
\end{figure}

\section{Method}
\subsection{Overview}
In semi‐supervised settings, we are given a small labeled set and a substantially larger unlabeled pool:
\[
  \mathcal{D}^l = \bigl\{(x_i^l, y_i^l)\bigr\}_{i=1}^{N_l}
  \quad\text{and}\quad
  \mathcal{D}^u = \bigl\{x_i^u\bigr\}_{i=1}^{N_u},
  \quad N_u \gg N_l,
\]
where each \(y_i^l\) consists of binary masks and class labels for every instance. Our goal is to distill knowledge from a large, pretrained VFM into a compact student \(f_{\theta_s}\), matching or surpassing the teacher’s accuracy with far fewer labels and compute. We propose \textbf{CAST}, a three-stage SSKD pipeline that hinges on two core innovations:
\ding{182} \underline{\emph{Contrastive Calibration.}}  We fine-tune a large VFM teacher via self-training, but rather than simple pseudo-labels we inject a pixel-wise contrastive head to sharpen mask boundaries. 
\ding{183} \underline{\emph{Debiased, Instance-Aware Sampling.}}  During both adaptation and distillation, we mine hard negatives via a joint mask-/class-probability embedding, focusing repulsion on informative inter-instance pairs tailored for instance segmentation.
These two ideas are then realized in three concise stages (see Fig.~\ref{fig:setup}):

\begin{enumerate}[leftmargin=6mm,nosep]
  \item \textbf{Teacher Adaptation.}  Self-train the VFM with pseudo-labels \emph{and} pixel-wise contrastive calibration to produce masks specialized to the target domain.  
  \item \textbf{Knowledge Transfer.}  Freeze this calibrated teacher and distill into a lightweight student under a unified loss that harmonizes ground truth, pseudo-label, and contrastive terms, guided by our debiased sampling.  
  \item \textbf{Student Refinement.}  Fine-tune the student on labeled data to remove residual pseudo-label bias.
\end{enumerate}
 
Sec.~\ref{sec:contrastive-loss} formalizes our instance-aware pixel-wise contrastive loss, which is used in both Teacher Adaptation and Knowledge Transfer to enforce intra-instance cohesion and inter-instance separation; Sec.~\ref{sec:pipeline} then details the three stages of the CAST pipeline.

\subsection{Pixel‐wise Contrastive Loss}
\label{sec:contrastive-loss}
Standard supervised and pseudo-label losses enforce correct mask predictions, ignoring pixel-level feature relationships which underutilize unlabeled data and amplify pseudo label noise. We therefore inject a self-supervised pixel-wise contrastive loss as an additional supervisory signal on both labeled and unlabeled images, sharpening feature discrimination and regularizing against noisy labels. 

Let $z^{\rm weak},z^{\rm strong}\in\ R^{B\times N\times D}$ 
be $\ell_2$‐normalized embeddings from two views of each image, where $B$ is the number of images in one mini batch, $N=h\times w$ the number of pixels, and $D$ the embedding dimension. For each pixel $p\in{1, 2, ..., N}$ and image index $b\in{1,...,B}$, the corresponding embedding vector is denoted as $z_{b,p}\in \mathbb{R}^D$.
We construct the positive pair by sampling the weak and strong embeddings for each pixel. The positive similarity between the two views is
\[
  s_{b,p}^+ = \langle z^{\rm weak}_{b,p},\,z^{\rm strong}_{b,p}\rangle/T.
\]
Negatives are sampled by our \emph{instance-aware} sampler (§\ref{par:debiased-sampling}), producing indices $\{(b',q_r)\}_{r=1}^R$ and corresponding similarities $s^-_{b,p,r}$.
\[
s^-_{b,p,r}
=\langle z^{\rm weak}_{b,p},\,z^{\rm strong}_{b',q_r}\rangle/T,
\quad r=1,\dots,R.
\]
The pixel‐wise contrastive loss is then the standard NT‐Xent over all anchors:
\[
\mathcal L_{\rm pxl}
= -\frac{1}{B\,N}
  \sum_{b=1}^B\sum_{p=1}^N
    \log\frac{\exp(s^+_{b,p})}
         {\exp(s^+_{b,p})+\sum_{r=1}^R\exp(s^-_{b,p,r})}.
\]
\noindent\textbf{Debiased Pixel‐Level Negative Sampling.}
\label{par:debiased-sampling}

To mine true inter‐instance pairs without quadratic cost, we derive a per pixel sampling distribution by fusing mask and class probabilities. Let
\(
M \in \mathbb{R}^{B \times K \times H \times W}
\), and \(
L \in \mathbb{R}^{B \times K \times (C+1)},
\)
be the model's mask and class logits respectively.  We first resize \(M\) to the feature resolution \((h\times w)\) and then normalize logits to probability distributions $P_m$ and $P_c$ via softmax along instance and class dimensions respectively.

For each pixel index \((b,p)\) to find the aggregated class vote, we compute Expected class distribution $F_c$. Further to avoid losing encoded instance ids over aggregation in expected class distribution we form a joint “pseudo probability” embedding by concatenation the mask distribution and class cues in a single vector which gives a richer embedding letting the contrastive head learn arbitrary interactions between mask and class. leading to pseudo probability map be $y[b,p]$.
\[
F_c[b,p,c] = \sum_{k=1}^K P_m[b,k,p]\;P_c[b,k,c],
\quad
y[b,p]
=\begin{bmatrix}
P_m[b,1:K,p]\\[6pt]
F_c[b,p,1:C+1]
\end{bmatrix}
\;\in\;\ R^{\,K + (C+1)}.
\]
We score any two pixels \((b,p)\neq(b',q)\) by $\tilde{y}$ being $\ell_2$‐normalized vector of pseudo probability map. 
\[
s^\mathrm{deb}\bigl((b,p),(b',q)\bigr)
= \max\bigl(0,\;1 - \langle\tilde y[b,p],\,\tilde y[b',q]\rangle\bigr),
\]
We draw $R$ negatives $\{q_r\}$ for each anchor $(b,p)$ by sampling proportional to $s^{\rm deb}$, and then plug these into the NT-Xent denominator of $\mathcal L_{\rm pxl}$.

\noindent\textbf{Theoretical Insight.}
To give a formal rationale for augmenting our pixel‐wise contrastive loss, we show that even under a mild negative sampling guarantee, each gradient step on our contrastive term provably increases the expected inter-instance margin.
\begin{assumption}[Negative Sampling Guarantee]\label{assump:sampling}
When sampling a negative under our instance aware scheme, the probability it originates from a different instance is at least $p>0.5$, where $p$ can be estimated empirically (see Sec.~\ref{emp}).
\end{assumption}
\begin{proposition}[Expected Margin Growth]\label{prop:margin-growth}
Under Assumption \ref{assump:sampling}, one gradient update on $\mathcal L_{\rm pxl}$ increases the expected inter-instance margin $\Delta_{\rm emp}$ by
\[
  \varepsilon = \Theta(p\,\lambda_{\rm pxl}) > 0.
\]
This expectation holds even when pseudo-labels are imperfect, provided negatives are sampled using our instance aware strategy.
\end{proposition}
In practice, raising $\lambda_{\rm pxl}$ enhances margin growth but also increases training cost. If $\lambda_{\rm pxl}$ is too large, it can overemphasize inter-instance separation at the expense of intra-instance cohesion. We validate this effect in Sec.~\ref{emp} and provide a proof sketch in Appendix C.

\subsection{CAST Framework}
\label{sec:pipeline}
We cast teacher adaptation, student distillation and student refinement as special cases of the same objective with three terms.  Let
\begingroup
\small
\[
  \mathcal{J}(\theta;\,\mathcal{D}^l,\mathcal{D}^u;\,\lambda_{\rm semi},\lambda_{\rm pxl})
  = 
  \underbrace{\frac{1}{N_l}\sum_{i=1}^{N_l}\ell\bigl(f_\theta(x_i^l),y_i^l\bigr)}_{\mathcal{L}_{\rm sup}}
  + \lambda_{\rm semi}\underbrace{\frac{1}{N_u}\sum_{j=1}^{N_u}\ell\bigl(f_\theta(x_j^u),\hat y_j^u\bigr)}_{\mathcal{L}_{\rm semi}}
  + \lambda_{\rm pxl}\,\mathcal{L}_{\rm pxl}\bigl(\theta;\,\mathcal{D}^l\cup\mathcal{D}^u\bigr),
\]
\endgroup
where $\mathcal{D}^u=\varnothing$ makes the middle term zero.

\noindent\textbf{Teacher adaptation.}
\label{sec:domain-adaptation}
Starting from pretrained weights $\theta_T^0$, we first fine‐tune on the labeled set \(\mathcal{D}^l\):
\[
     \theta_T'
     = \arg\min_{\theta}\;
     \mathcal{J}\bigl(\theta;\,\mathcal{D}^l,\varnothing;\,0,\lambda_{\rm pxl}\bigr).
   \]
We then generate pseudo‐labels
\(\hat y_j^u = f_{\theta_T'}(x_j^u)\), reset to \(\theta_T^0\)and fine-tune on $\mathcal{D}^l\cup\{\!(x_j^u,\hat y_j^u)\!\}$:
\[
     \theta_T''
     = \arg\min_{\theta}\;
     \mathcal{J}\bigl(\theta;\,\mathcal{D}^l,\mathcal{D}^u;\,1,\lambda_{\rm pxl}\bigr).
   \]
This two‐step contrastive calibration yields a specialized teacher whose pseudo-labels are both accurate and spatially consistent for the target domain.

\noindent\textbf{Knowledge transfer.}
\label{sec:dist}
With calibrated teacher $\theta_T''$ frozen, student $\theta_s$ is trained via the unified objective:
\begin{equation}
\label{eq:student-distill}
  \theta_s^*
  = \arg\min_{\theta_s}\;
  \mathcal{J}\bigl(\theta_s;\,\mathcal{D}^l,\mathcal{D}^u;\,\lambda_{\rm semi},\lambda_{\rm pxl}\bigr).
\end{equation}
Here, $\mathcal{L}_{\rm sup}$ enforces ground truth supervision on $\mathcal{D}^l$, $\mathcal{L}_{\rm semi}$ distills pseudo-labels from $\mathcal{D}^u$, and $\mathcal{L}_{\rm pxl}$ imposes our pixel-wise contrastive regularizer across both sets. The coefficients $\lambda_{\rm semi}$ and $\lambda_{\rm pxl}$ balance signals, guiding the student to approach teacher's accuracy with far fewer parameters.

\noindent\textbf{Student Refinement.}
\label{sec:student-refinement}
Although joint distillation yields a strong initialization, residual pseudo-label noise and contrastive pretext tasks can introduce bias. As a final step, we fine-tune the student on labeled data alone:
\[
  \theta_s^{\dagger}
  = \arg\min_{\theta_s^*}\;
  \mathcal{J}\bigl(\theta_s^*;\,\mathcal{D}^l,\varnothing;\,0,0\bigr),
\]
This pass removes pseudo-label drift and sharpens decision boundaries for in‐domain data.  

\section{Experiments}
\subsection{Experimental Protocol}
\noindent\textbf{Datasets.}  
We evaluate CAST on two standard instance segmentation benchmarks: \textbf{Cityscapes}~\cite{Cordts2016Cityscapes} contains 2,975 training, 500 validation images of urban street scenes, annotated with 19 semantic categories (8 “thing” classes and 11 “stuff” classes). \textbf{ADE20K}~\cite{zhou2019semantic} comprises 20,210 training and 2,000 validation images spanning diverse indoor and outdoor environments, annotated with 150 semantic categories (100 “thing” and 50 “stuff” classes).

\noindent\textbf{Implementation Details.}
All experiments were conducted on Ubuntu~22.04 with Python~3.10 and PyTorch~2.6.0 (CUDA~12.6). Teacher adaptation runs were executed on 2$\times$NVIDIA~A100 GPUs, while student training runs used 2$\times$NVIDIA GeForce RTX 4090 GPUs. As a reference, a single fine-tuning run of the teacher (Grounding-DINO) on the supervised Cityscapes split required $\approx3.5$~GPU hours; a single student training run for this dataset took $\approx17$~GPU hours. 


\noindent\textbf{Teacher and Student Architectures.}
Our teacher is a fused ensemble of Grounding-DINO-Large~\cite{liu2024grounding} and SAM2-L~\cite{ravi2024sam}. Since the SOTA model of Grounding-DINO is closed-source, we use its open-source counterpart mm-Grounding-DINO~\cite{zhao2024open}. For the student, we pair a DINOv2-S encoder~\cite{oquab2023dinov2} with a DPT-S decoder head~\cite{ranftl2021vision}, followed by a lightweight transformer decoder module in the spirit of Mask2Former~\cite{cheng2022masked}. Our choice of the DINOv2+DPT backbone is motivated by the recent successes of “Depth AnythingV2” in monocular depth estimation~\cite{yang2024depth} and UniMatchV2 in semantic segmentation~\cite{yang2025unimatch}, and aims to facilitate future multimodal fusion work. We evaluate the impact of different student designs in Sec.~\ref{encoder-decoder}, and defer the complete optimizer, learning rate schedules, and other hyperparameters to Appendix~B.

\subsection{Main Results}
\label{sec:results}
We evaluate a range of knowledge distillation (KD) strategies, ranging from purely supervised to state‐of‐the‐art semi‐supervised baselines, and benchmark them against our CAST pipeline. Table~\ref{tab:results_both} reports maskAP and maskAP$_{50}$ on Cityscapes and ADE20K. In the teacher adaptation stage (568M parameters), adding our pixel‐level contrastive loss boosts Cityscapes maskAP from 29.7 to 30.5 (+0.8) and maskAP$_{50}$ from 54.9 to 56.6 (+1.7); on ADE20K, maskAP rises from 14.6 to 15.2 (+0.6) and maskAP$_{50}$ from 23.6 to 24.5 (+0.9). These improvements confirm that pixel‐wise supervision sharply improves feature discrimination and reduces pseudo-label noise. 

\begin{wrapfigure}[13]{r}[0pt]{0.5\textwidth}
  \centering
  \vspace{-10pt}
  \includegraphics[width=\linewidth]{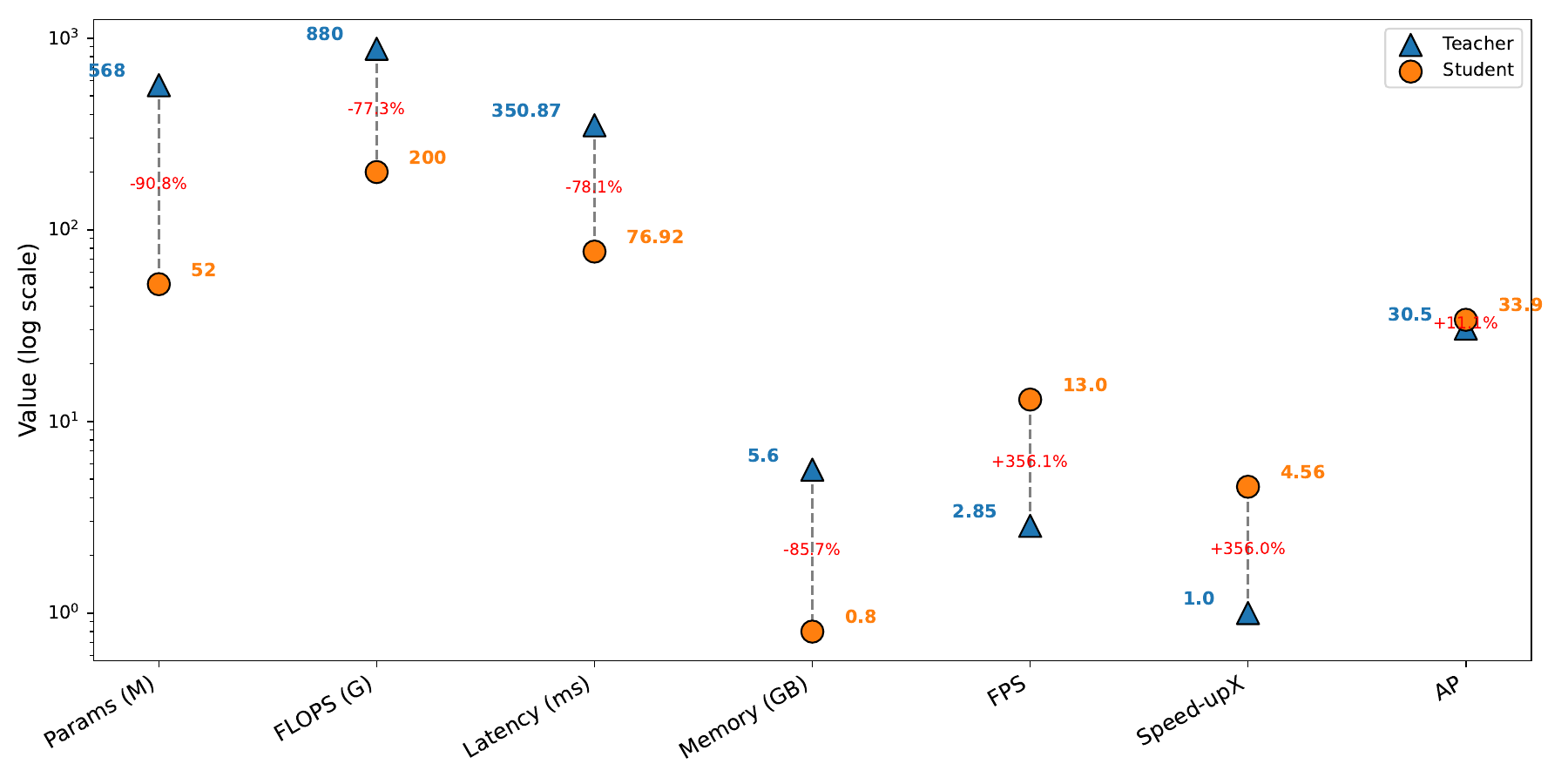}
  \vspace{-3mm}
  \caption{Efficiency comparison (log scale).
  }
  \label{fig:scatter_chart}
\end{wrapfigure}

In the student distillation stage, our 52M-parameter student (9\% of the composite teacher model) achieves 32.2 maskAP and 56.5 maskAP$_{50}$ on Cityscapes with pixel-level loss, outperforming prior SOTA SSKD models. After fine-tuning, the student reaches 33.9 maskAP (+3.4 over the best teacher) and 58.7 maskAP$_{50}$. On ADE20K, it attains 16.1 maskAP and 27.4 maskAP$_{50}$ in the semi-supervised setting, and improves further to 16.7 maskAP (+1.5) and 28.0 maskAP$_{50}$ after fine-tuning, underscoring CAST’s robustness across benchmarks. Additional ablations under varied label splits are presented in Section~\ref{split}. To compare efficiency, Figure~\ref{fig:scatter_chart} plots key pipeline efficiency metric on a logarithmic scale for both teacher and student models.

\begin{table*}[ht]
  \centering
  \caption{\textbf{Main results on Cityscapes and ADE20K} with 10\% labeled data.
  We report teacher adaptation (568M) and student distillation (52M).
  * denotes adapted methods. Rows in gray are ours.}
  \label{tab:results_both}
  \setlength\tabcolsep{6pt}
  \begin{adjustbox}{width=\textwidth}
  \begin{tabular}{l l S[table-format=2.1, ] S[table-format=2.1] S[table-format=2.1] S[table-format=2.1]}
    \toprule
    \textbf{Method} & \textbf{Data Regime}
      & \multicolumn{2}{c}{\textbf{Cityscapes}}
      & \multicolumn{2}{c}{\textbf{ADE20K}} \\
    \cmidrule(lr){3-4}\cmidrule(lr){5-6}
    & & {maskAP} & {maskAP$_{50}$} & {maskAP} & {maskAP$_{50}$} \\
    \midrule[0.75pt]

    \multicolumn{6}{c}{\textit{Teacher Adaptation}} \\
    \cmidrule(lr){1-6}
    Zero-shot VFM & None (pretrained) & 22.0 & 42.3 &  8.1 & 18.2 \\
    Supervised fine-tuning & Labeled only & 28.7 & 53.4 & 14.2 & 23.5 \\
    Self-training*~\cite{xie2020self} & Labeled+Unlabeled & 29.7 & {\underline {54.9}} & 14.6 & 23.6 \\
    Unbiased Teacher*~\cite{liu2021unbiased} & Labeled+Unlabeled & {\underline {29.8}} & {\underline {54.9}} & {\underline {14.8}} & {\underline {23.7}} \\
    CAST (teacher adaptation) & Labeled+Unlabeled & {\bfseries 30.5} & {\bfseries 56.6} & {\bfseries 15.2} & {\bfseries 24.5} \\
    \midrule[0.75pt]

    \multicolumn{6}{c}{\textit{Student Distillation}} \\
    \cmidrule(lr){1-6}
    Supervised fine-tuning & Labeled only & 21.1 & 38.7 & 13.9 & 24.2 \\
    PAIS~\cite{hu2023pseudo} & Labeled+Unlabeled & 22.9 & 44.9 & 10.3 & 18.3 \\
    Guided dist.~\cite{berrada2024guided} & Labeled+Unlabeled & 30.8 & 52.9 & 14.2 & 23.8 \\
    Vemulapalli et al.*~\cite{Vemulapalli2024KD} & Unlabeled only & 24.4 & 45.6 & 5.1 & 9.3 \\
    CAST (knowledge transfer) & Labeled+Unlabeled & {\underline {32.2}} &  {\underline {56.5}} &  {\underline {16.1}} &  {\underline {27.4}} \\
    CAST (student refinement) & Labeled only & {\bfseries 33.9} & {\bfseries 58.7} & {\bfseries 16.7} & {\bfseries 28.0} \\
    \bottomrule
  \end{tabular}
  \end{adjustbox}
\end{table*}

\subsection{Empirical Validation}
\label{emp}
\begin{figure}[ht]
  \centering
  \begin{subfigure}[b]{0.32\textwidth}
    \includegraphics[width=\textwidth]{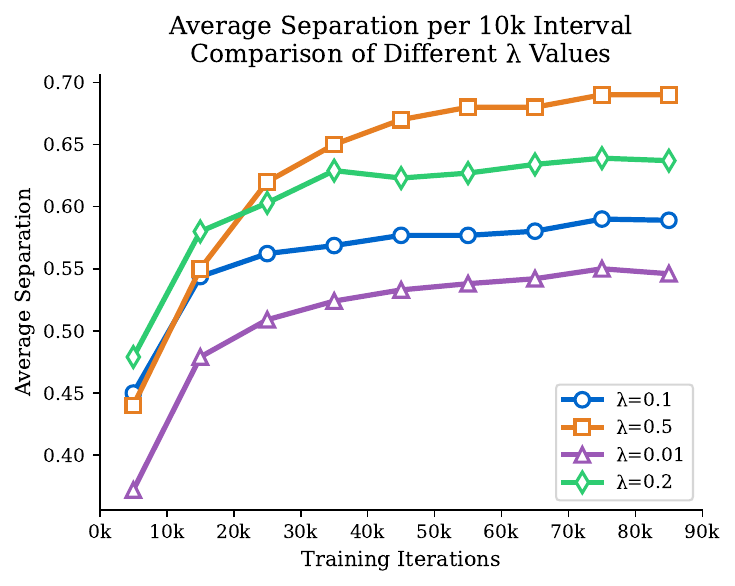}
    \caption{Instance Separation}
    \label{fig:sep}
  \end{subfigure}\hfill
  \begin{subfigure}[b]{0.32\textwidth}
    \includegraphics[width=\textwidth]{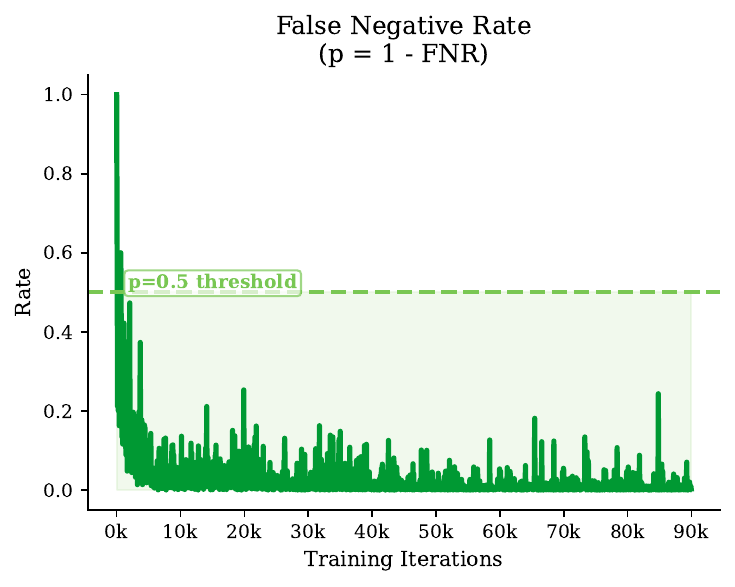}
    \caption{$\mathrm{FNR}$}
    \label{fig:fnr}
  \end{subfigure}\hfill
  \begin{subfigure}[b]{0.32\textwidth}
    \includegraphics[width=\textwidth]{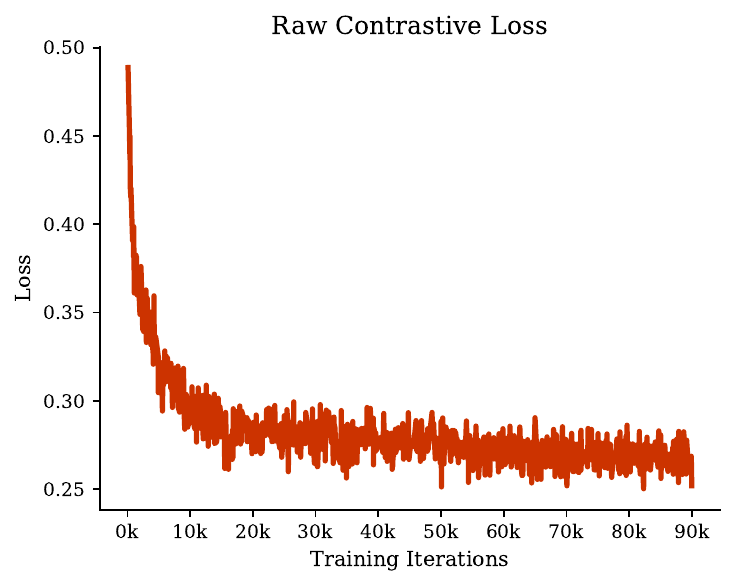}
    \caption{Contrastive Loss}
    \label{fig:loss}
  \end{subfigure}
  \caption{%
    \textbf{(Left)} Empirical margin (NegMean–PosMean) every 10k iterations for various $\lambda_{\rm pxl}$.  
    \textbf{(Center)} False negative rate ($\mathrm{FNR}$) for $\lambda_{\rm pxl}=0.1$, dashed at $p=0.5$.  
    \textbf{(Right)} Contrastive loss for $\lambda_{\rm pxl}=0.1$.}
  \label{fig:metrics}
\end{figure}

We validate Proposition~\ref{prop:margin-growth} by monitoring the false negative rate ($\mathrm{FNR}$), the fraction of sampled negatives that actually belong to the same instance, and the empirical margin 
\[
  \Delta_{\rm emp} = \mathrm{NegMean} - \mathrm{PosMean}.
\]
Defining $p = 1 - \mathrm{FNR}$ as the success probability of sampling a true negative, Figure~\ref{fig:metrics} shows:
the empirical margin every 10 k iterations for $\lambda_{\rm pxl}\in\{0.01,0.05,0.1,0.2\}$ (left), the raw contrastive loss for $\lambda_{\rm pxl}=0.1$ (center), and the false negative rate for $\lambda_{\rm pxl}=0.1$ (right, dashed at $p=0.5$). Throughout training we observe $p>0.9$ and a linear increase of $\Delta_{\rm emp}$ with $\lambda_{\rm pxl}$, in agreement with Proposition~\ref{prop:margin-growth}.

\subsection{Ablation Studies}
We perform a series of ablation experiments to isolate the contributions of each component in the CAST pipeline. These include analyses of loss functions, training stages, negative sampling strategies, hyperparameters, and student architecture choices.

\noindent\textbf{Impact of Loss Components.}
During distillation, the objective combines three terms: supervised loss (\(\mathcal{L}_{\text{sup}}\)), semi-supervised pseudo-label loss (\(\mathcal{L}_{\text{semi}}\)), and pixel-level self-supervised contrastive loss (\(\mathcal{L}_{\text{pxl}}\)). 
Table~\ref{tab:loss_ablation} shows that adding \(\mathcal{L}_{\text{semi}}\) improves student performance from 21.1 to 30.7 maskAP, while further including \(\mathcal{L}_{\text{pxl}}\) yields the best result of 32.2 maskAP, confirming complementary benefit.

\begin{table}[htbp]
  \centering
  \caption{Ablations on Cityscapes (10\% labels). Left: effect of loss terms. Right: effect of CAST stages.}
  \vspace{0.5em}
  \begin{minipage}{0.48\linewidth}
    \centering
    \resizebox{\linewidth}{!}{
    \begin{tabular}{lccc cc}
      \toprule
      \textbf{Method} & $\mathcal{L}_{\rm sup}$ & $\mathcal{L}_{\rm semi}$ & $\mathcal{L}_{\rm pxl}$ & Teacher & Student \\
      \midrule
      (a) Sup. only     & \checkmark &   &   & 28.7 & 21.1 \\
      (b) + Pseudo      & \checkmark & \checkmark &   & 29.7 & 30.7 \\
      (c) + Pixel loss  & \checkmark &   & \checkmark & 29.6 & 27.5 \\
      (d) (b)+(c)       & \checkmark & \checkmark & \checkmark & \bf 30.5 & \bf 32.2 \\
      \bottomrule
    \end{tabular}}
    \vspace{-0.5em}
    \subcaption{Loss ablation}
    \label{tab:loss_ablation}
  \end{minipage}
  \hfill
  \begin{minipage}{0.48\linewidth}
    \centering
    \resizebox{\linewidth}{!}{
    \begin{tabular}{lccc c}
      \toprule
      \textbf{Variant} & Teacher Adapt. & Distill. & Student FT & maskAP \\
      \midrule
      Full CAST             & \checkmark & \checkmark & \checkmark & 33.9 \\
      No Student FT         & \checkmark & \checkmark &            & 32.2 \\
      No Teacher Adapt.     &            & \checkmark & \checkmark & 25.7 \\
      Distillation Only     &            & \checkmark &            & 23.8 \\
      No Distill. (Sup.)    &            &            & \checkmark & 21.1 \\
      \bottomrule
    \end{tabular}}
    \vspace{-0.5em}
    \subcaption{Stage ablation}
    \label{tab:stage_ablation}
  \end{minipage}
\end{table}

\noindent\textbf{Impact of Training Stages.}
Beyond the contribution of individual loss terms, we further ablate each stage of CAST to justify their necessity. 
Table~\ref{tab:stage_ablation} shows results on Cityscapes (10\% labels), where we drop exactly one stage at a time.

\noindent The supervised baseline achieves 21.1 maskAP. Adding distillation alone improves this to 23.8 (+2.7), and further adding student fine-tuning raises it to 32.2 (+8.4). 
Without teacher adaptation, performance drops to 25.7, underscoring the need to align the teacher with the target domain. 
The full three-stage CAST pipeline achieves best result of 33.9 maskAP, a +12.8 improvement over baseline.

\noindent\textbf{Ablation of Negative Sampling via Various Probability Maps.}
To validate our negative sampling strategy in the pixel-level contrastive loss, Table~\ref{tab:mask_class_fusion} compares four sampling methods: \textbf{Uniform:} negatives sampled uniformly across the image; \textbf{Mask-Only:} The probability map is derived solely from mask predictions, with class probabilities assumed to be uniform. \textbf{Class-Only:} The map is generated only from class predictions, assuming a uniform spatial distribution for the mask.  \textbf{Fusion:} Combining both mask and class predictions. The fusion strategy achieves the best results, with 32.2 maskAP and 56.5 AP$_{50}$.

\begin{table}[htbp]
  \centering
  \small
  \setlength\tabcolsep{4pt}
  \begin{minipage}[t]{0.40\linewidth}
    \centering
    \subcaption{Negative Sampling Strategies}
    \label{tab:mask_class_fusion}
    \begin{tabular}{lSS}
      \toprule
      Method      & {maskAP (\%)} & {maskAP$_{50}$ (\%)} \\
      \midrule
      Uniform      & 29.4   & 50.2            \\
      Mask‐Only    & 30.6          & 55.0           \\
      Class‐Only   & 31.1          & 55.3           \\
      Fusion       & \bfseries {32.2}& \bfseries {56.5}\\
      \bottomrule
    \end{tabular}
  \end{minipage}%
  \hfill
  \begin{minipage}[t]{0.60\linewidth}
    \centering
    \subcaption{Schematic of Sampling Distributions}
    \vspace{-0.1\baselineskip}
\label{fig:sampling_pdf}
\includegraphics[%
  width=0.6\linewidth,%
  height=2.5cm,%
  keepaspectratio=false%
]{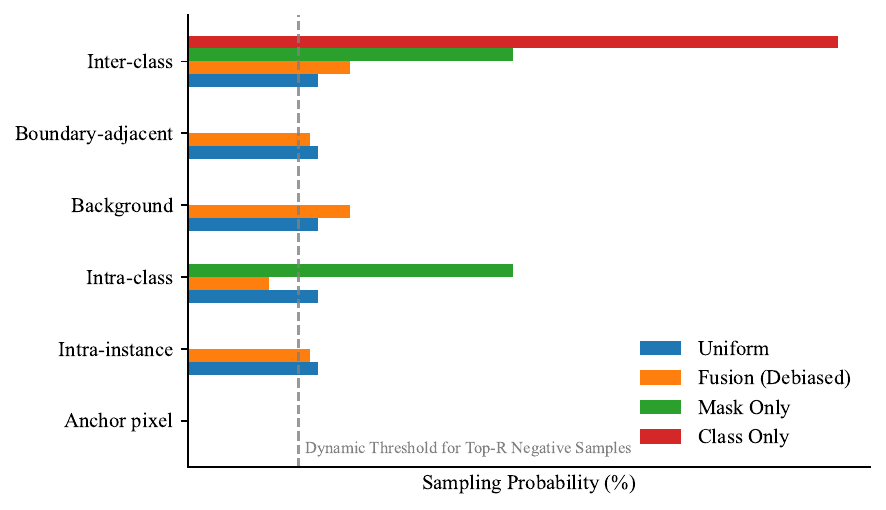}
  \end{minipage}
  \caption{%
    \textbf{Ablation of Negative Sampling Strategies on Cityscapes.}
    (\subref{tab:mask_class_fusion}) Quantitative results for uniform, mask‐only, class‐only, and fusion samplers (maskAP and maskAP$_{50}$). 
    (\subref{fig:sampling_pdf}) Schematic sketch of the corresponding pixel‐level sampling probability distributions.}
\label{tab:sampling_ablation}
\end{table}

\noindent\textbf{Hyperparameter Sensitivity.}
\label{sec:ablate-hyperparam}
We evaluate CAST’s sensitivity to three key hyperparameters on Cityscapes: contrastive weight $\lambda_{\rm pxl}$, negatives per anchor $K$, and temperature $T$, by measuring both teacher and student maskAP (\%) and maskAP$_{50}$ (\%). Table~\ref{tab:hyperparam} reports the full sweep.  We find that $\lambda_{\rm pxl}=0.2$ and $T=0.2$ consistently maximize performance. For the number of negatives, $K=256$ offers the best trade-off: although $K=512$ yields a slight increase in teacher maskAP (30.9 vs.\ 30.5) and maskAP$_{50}$ (57.1 vs.\ 56.6), and comparable student metrics, the marginal gains saturate relative to the increased sampling cost. Therefore, we adopt $K=256$ throughout.
\begin{table}[!htbp]
  \centering
  \small
\caption{\textbf{Hyperparameter Ablation on Cityscapes.}}
  \label{tab:hyperparam}
  \resizebox{\linewidth}{!}{%
  \begin{tabular}{ll
                  SSSSS   
                  SSS     
                  SSS     
                  }
    \toprule
    \multirow{2}{*}{Model} 
      & \multirow{2}{*}{Metric}
      & \multicolumn{5}{c}{Contrastive Loss Weight($\lambda_{\rm pxl}$)}
      & \multicolumn{3}{c}{Negative Sanples per Anchor($K$)}
      & \multicolumn{3}{c}{Temperature($T$)} \\
    \cmidrule(lr){3-7} \cmidrule(lr){8-10} \cmidrule(lr){11-13}
      & 
      & {0} & {0.01} & {0.1} & {\bfseries{0.2}} & {0.5}
      & {128} & \bfseries{{256}} & {512}
      & {0.1} & \bfseries{{0.2}} & {0.4} \\
    \midrule
    \multirow{2}{*}{Teacher}
      & AP    & 29.7 & 29.9 & 30.2 & \bfseries 30.5 & 30.1
              & 30.4 & \bfseries 30.5 & 30.9
              & 30.1 & \bfseries 30.5 & 29.8 \\
      & AP$_{50}$ 
              & 55.3 & 55.7 & 56.1 & \bfseries 56.6 & 56.1
              & 56.3 & \bfseries 56.6 & 57.1
              & 55.9 & \bfseries 56.6 & 55.3 \\
    \addlinespace
    \multirow{2}{*}{Student}
      & AP    & 30.7 & 30.8 & 32.1 & \bfseries 32.2 & 30.9
              & 29.8 & \bfseries 32.2 & 32.1
              & 31.9 & \bfseries 32.2 & 31.7 \\
      & AP$_{50}$
              & 54.9 & 55.2 & 56.2 & \bfseries 56.5 & 55.7
              & 55.3 & \bfseries 56.5 & 56.6
              & 56.0 & \bfseries 56.5 & 55.8 \\
    \bottomrule
  \end{tabular}
  }
\end{table}

\noindent\textbf{Student Architecture Variants.}
\label{encoder-decoder}
We evaluate two design axes for the student model under CAST distillation protocol: (i) the encoder backbone (with a fixed DPT decoder), and (ii) the decoder head (with a fixed DINOv2-S encoder). Table~\ref{tab:arch_ablation} reports accuracy along with parameter counts, on the Cityscapes validation set. The combination of DINOv2-S encoder and DPT head achieves the best accuracy with a compact footprint.

\begin{table}[!htbp]
  \centering
  \small
  \caption{\textbf{Architecture Ablations on Cityscapes.}  
    (\subref{tab:encoder_ablation}) Encoder backbone (fixed DPT decoder).  
    (\subref{tab:decoder_ablation}) Decoder head (fixed DINOv2-S encoder).}
  \label{tab:arch_ablation}
  \begin{subtable}[t]{0.48\linewidth}
    \centering
  \renewcommand{\arraystretch}{0.7}
    \caption{Encoder Backbone}
    \label{tab:encoder_ablation}
    \begin{tabular}{lccc}
      \toprule
      Encoder    & maskAP & maskAP$_{50}$ & Params (M) \\
      \midrule
      ResNet50   & 25.5   & 49.3      & 24         \\
      SAM2-S     & 22.1   & 39.2      & 35         \\
      DINOv2-S   & \bfseries 30.7 & \bfseries 54.9 & 22 \\
      \bottomrule
    \end{tabular}
  \end{subtable}
  \hfill
  \begin{subtable}[t]{0.48\linewidth}
    \centering
    \caption{Decoder Head}
    \label{tab:decoder_ablation}
    \begin{tabular}{lccc}
      \toprule
      Decoder    & maskAP & maskAP$_{50}$ & Params (M) \\
      \midrule
      FPN        & 28.9   & 52.4      & 18         \\
      DPT        & \bfseries 30.7 & \bfseries 54.9 & 22 \\
      \bottomrule
    \end{tabular}
  \end{subtable}
\end{table}

\begin{figure*}[!htpb]
\begin{center}
   \includegraphics[width=1\linewidth]{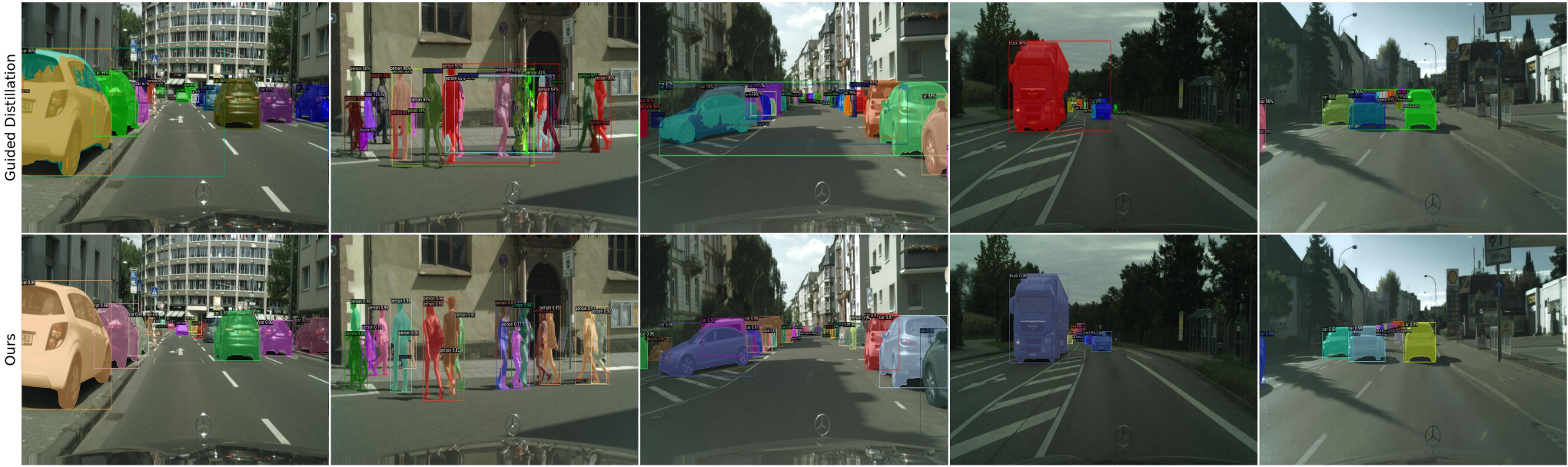}
   \end{center}
   \caption{%
    \textbf{Qualitative results on Cityscapes.} Guided dist.~\cite{berrada2024guided} (top) vs.\ CAST (bottom).}
\label{fig:qual}
\end{figure*}

\noindent\textbf{Scalability with Labeled Fractions.}
\label{split}
We evaluate CAST under different fractions of labeled data to assess scalability in semi-supervised settings. Following the protocol in~\cite{berrada2024guided}, we train with 5\%, 10\%, and 30\% labeled splits of Cityscapes. As shown in Table~\ref{tab:scalability}, CAST consistently outperforms prior methods across all fractions. At 5\% labels, CAST achieves 30.7 AP, far exceeding PAIS (18.0) and Guided Distillation (23.0). At 30\% labels, CAST reaches 40.4 AP, surpassing the strongest baseline (37.8 from S$^4$M) by +2.6 AP. These results demonstrate that CAST remains effective under scarce supervision while scaling gracefully with additional labeled data.

\begin{table}[htbp]
  \centering
  \caption{\textbf{Scalability across label fractions on Cityscapes.} Results with 5\%, 10\%, and 30\% labeled data.}
  \label{tab:scalability}
  \resizebox{0.85\linewidth}{!}{
  \begin{tabular}{lcccccc}
    \toprule
    \textbf{Dataset Fraction} & Teacher Adapt. & Distillation & CAST (student) & PAIS~\cite{hu2023pseudo} & Guided dist.~\cite{berrada2024guided} & S$^4$M~\cite{yoon2025s} \\
    \midrule
    5\%   & 29.4 & 29.2 & \textbf{30.7} & 18.0 & 23.0 & 30.1 \\
    10\%  & 30.5 & 32.2 & \textbf{33.9} & 22.9 & 30.8 & 33.3 \\
    30\%  & 33.3 & 38.5 & \textbf{40.4} & 32.8 & 35.6 & 37.8 \\
    \bottomrule
  \end{tabular}
  }
\end{table}

\noindent Additional ablations, including teacher adaptation variants, loss formulations, sampling scope, and backbone comparisons, are provided in the supplementary material (Section~E).

\section{Conclusions}
\setlength{\intextsep}{0.5\baselineskip}   
\begin{wrapfigure}[15]{r}[0pt]{0.4\textwidth}
  \centering
  \includegraphics[width=\linewidth]{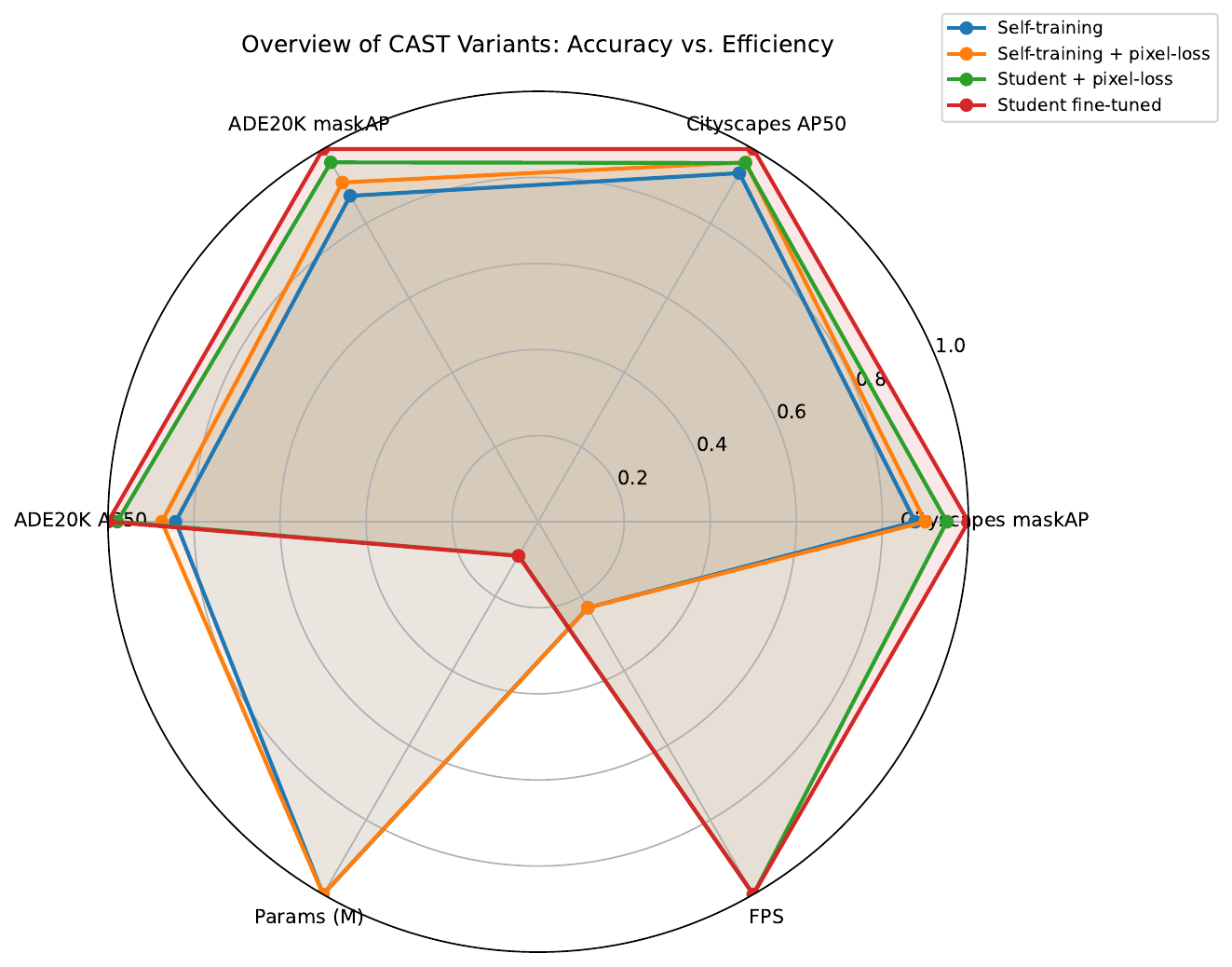}
  \caption{Performance–complexity radar chart (normalized).
  }
  \label{fig:radar_chart}
\end{wrapfigure}

We have introduced CAST, a rigorously designed SSKD pipeline that fuses self‐training, instance-aware pixel‐wise contrastive learning, and final supervised finetuning to compress large VFMs into compact student experts with comparable performance. Empirically, our $\approx11\times$ smaller student exceeds its adapted teacher by +3.4 maskAP in Cityscapes and +1.5 maskAP in ADE20K, while cutting compute and parameter counts demonstrating that dense contrastive supervision can unlock substantial gains in low‐label regimes. Our theoretical analysis further guarantees that our negative sampling scheme provably increases inter‐instance margins under mild assumptions. Looking forward, streamlining CAST into a single unified objective, extending its evaluation to diverse domains, and integrating uncertainty quantification will be critical steps toward safe, equitable, and broadly deployable segmentation solutions.

\begin{figure*}[!htpb]
\begin{center}
   \includegraphics[width=0.97\linewidth]{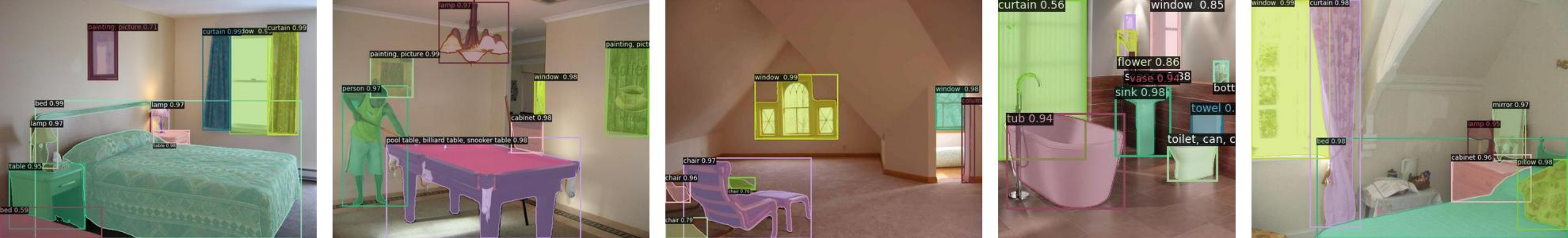}
   \end{center}
   \caption{%
    \textbf{Qualitative results on ADE20K.}
    }
\label{fig:qual2}
\end{figure*}

\begin{figure*}[htpb]
\begin{center}
   \includegraphics[width=1\linewidth]{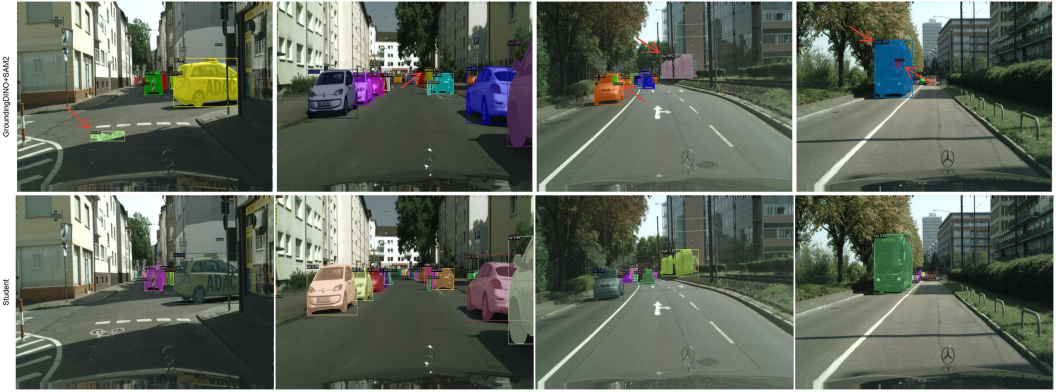}
   \end{center}
\caption{\textbf{Qualitative bias reduction in stage‐wise distillation.} 
Top row: pseudo‐labels generated by the pretrained teacher. 
Bottom row: student predictions after distillation and refinement, demonstrating reduced pseudo‐label bias and sharper instance boundaries.}
\label{fig:qual3}
\end{figure*}

\newpage

\section*{Ethics Statement}
This work does not involve human subjects, private data, or sensitive content. All datasets used are publicly available. Portions of the manuscript were polished using large language model (LLM) for clarity; this use was limited to text editing and did not affect the research process, experiments, or results.

\section*{Reproducibility Statement}
We provide detailed descriptions of datasets, model architectures, and training procedures to ensure reproducibility. All datasets (Cityscapes, ADE20K) are publicly available, and the 5\%, 10\%, and 30\% labeled splits follow established protocols (Section~\ref{split}). Implementation details, including hyperparameters, software environment, and GPU usage, are reported in Section~\ref{sec:results}, with extensive ablations and sensitivity analyses in Appendix~E. To facilitate further research, we will release our code upon the completion of the anonymous review process.

{
    \small
    \bibliographystyle{ieeenat_fullname}
    \bibliography{main}
}

\appendix

\section*{Supplementary Material}
\label{supplementary}

This document provides additional details to support the main paper, including dataset statistics, full hyperparameter settings, formal proof, extended training protocols, and additional ablation studies. 

\section{Dataset Splits}
\label{sec:datasets}

Table~\ref{tab:datasets} summarizes the datasets used in our experiments. We use a 10\% labelled split of Cityscapes’ 2\,975 training images (298 labeled / 2\,677 unlabeled) and a stratified 20\% split of ADE20K’s 20\,210 training images (1\,000 labeled / 2\,537 unlabeled). Standard validation sets are retained (500 images for Cityscapes, 2\,000 for ADE20K). Exact image‐ID lists will be released with our code.

\begin{table}[ht]
  \centering
  \small
  \caption{Semi-supervised splits used in our experiments.}
  \label{tab:datasets}
  \begin{tabular}{lccc}
    \toprule
    Dataset          & \# Classes & Labeled / Unlabeled & Validation \\
    \midrule
    Cityscapes       & 8          & 298 / 2\,677        & 500        \\
    ADE20K           & 100        & 1\,000 / 2\,537     & 2\,000      \\
    \bottomrule
  \end{tabular}
\end{table}

\section{Hyperparameters}
\label{sec:hyperparams}

Key teacher and student hyperparameters are summarized in Table~\ref{tab:hyperparams}.

\begin{table}[ht]
  \centering
  \small
  \caption{Hyperparameter Settings}
  \label{tab:hyperparams}
  \begin{tabular}{lcc}
    \toprule
    \textbf{Parameter}                  & \textbf{Teacher}                       & \textbf{Student}                                          \\
    \midrule
    Learning rate                       & $5.0\times10^{-5}$                     & Encoder: $5.0\times10^{-6}$; Decoder: $5.0\times10^{-5}$ \\
    Scheduler                           & Multi-step (milestones at 0.9, 0.95)   & PolyLR (power 0.9)                                        \\
    Batch size                          & 4                                      & 8                                                         \\
    \midrule
    Weight decay   & 0.01 & 0.05  \\
    Contrastive loss weight    & 0.2   & 0.2    \\
    Pseudo-label threshold    & 0.3   & 0.3    \\
    Dropout rate         & —       & 0.1   \\
    Gradient clipping    & —    & $\ell_2$ norm 0.1   \\
    \midrule
    Optimizer    & \multicolumn{2}{c}{AdamW ($\beta 1$=0.9, $\beta 2$=0.999)}      \\
    Augmentations     & \multicolumn{2}{c}{Weak: flip, resize; Strong: random resized crop, jitter, grayscale, blur, } \\
    Loss weights (mask / class)         & \multicolumn{2}{c}{5 / 2}                                                \\
    \bottomrule
  \end{tabular}
\end{table}

\section{Proof Sketch of Proposition~3.1}
\label{sec:prof-sketch}

\begin{proof}[Proof Sketch]
Let \(z_a\), \(z^+\) and \(\{z^-_r\}_{r=1}^R\) be the unit norm embeddings of an anchor pixel, its positive, and \(R\) negatives.  Define
\[
s^+ = \langle z_a,\,z^+\rangle,
\qquad
s^-_r = \langle z_a,\,z^-_r\rangle,
\]
and the pixel-wise contrastive loss
\[
\ell(z_a)
= -\log\frac{\exp(s^+)}{\exp(s^+)+\sum_{r=1}^R\exp(s^-_r)}.
\]
Let
\[
Z = \exp(s^+)+\sum_{r=1}^R\exp(s^-_r),
\qquad
\alpha_r = \frac{\exp(s^-_r)}{Z}.
\]
A straightforward gradient computation gives
\[
\nabla_{z_a}\ell
= \sum_{r=1}^R \alpha_r\,(z^-_r - z^+).
\]
Applying one gradient descent step with step size \(\lambda_{\rm pxl}\):
\[
z_a' = z_a - \lambda_{\rm pxl}\,\nabla_{z_a}\ell
\;=\;
z_a + \lambda_{\rm pxl}\sum_{r=1}^R \alpha_r\,(z^+ - z^-_r).
\]
For a randomly chosen negative \(z^-\),
\begin{align*}
\Delta s^+
&= \langle z_a' - z_a,\,z^+\rangle
= \lambda_{\rm pxl}\sum_{r=1}^R \alpha_r\bigl(1 - \langle z^-_r,\,z^+\rangle\bigr),\\
\Delta s^-
&= \langle z_a' - z_a,\,z^-\rangle
= \lambda_{\rm pxl}\sum_{r=1}^R \alpha_r\bigl(\langle z^+,\,z^-\rangle - \langle z^-_r,\,z^-\rangle\bigr).
\end{align*}

By Assumption~3.1, each negative embedding \(z^-_r\) is inter-instance with probability \(p\), in which case
\(\langle z^-_r,\,z^+\rangle\approx0\), and intra-instance with probability \(1-p\), in which case
\(\langle z^-_r,\,z^+\rangle\approx1\).  Hence
\[
  \EE\bigl[1 - \langle z^-_r,\,z^+\rangle\bigr]
  = p\cdot1 + (1-p)\cdot0
  = p,
\]
and since \(\sum_{r=1}^R\alpha_r=1\), it follows that
\[
  \EE[\Delta s^+]
  = \lambda_{\rm pxl}\sum_{r=1}^R\alpha_r\,
    \EE\bigl[1 - \langle z^-_r,\,z^+\rangle\bigr]
  = p\,\lambda_{\rm pxl}.
\]
Meanwhile, every term in \(\Delta s^-\) involves an inter-instance inner product, either
\(\langle z^+,z^-\rangle\) or \(\langle z^-_r,z^-\rangle\) each of which vanishes in expectation, so
\(\EE[\Delta s^-]\approx0\).  Therefore
\[
  \EE[\Delta s^+ - \Delta s^-]
  = p\,\lambda_{\rm pxl} - 0
  = \Theta\!\bigl(p\,\lambda_{\rm pxl}\bigr)
  = \varepsilon>0,
\]
i.e.\ one update on \(\mathcal L_{\rm pxl}\) increases the expected inter-instance margin by \(\varepsilon\).
\end{proof}

\begin{remark}[Why \(\langle z^+, z^-\rangle \approx 0\) holds]
Under the InfoNCE objective (§3.2), the normalized weights for negative pairs,
\(\alpha_r = \frac{e^{s^-_r}}{e^{s^+} + \sum_{r}e^{s^-_r}},\)
vanish at convergence, i.e.\ \(\alpha_r \approx 0\).  Moreover, in high dimensional embeddings, random unit vectors have inner products concentrating near zero, and contrastive training further pushes these negative similarities into a tight, small magnitude distribution~\cite{chen2020simple}.  Thus it is reasonable to approximate \(\langle z^+,z^-\rangle\approx0\) up to \(O(1/\sqrt{D})\) fluctuations.
\end{remark}

\section{More Training Details}
\label{sec:training-details}
All teacher models are fine-tuned using 1k iterations on labeled set, followed by 5k iterations in a self-training stage with pseudo-labels. For student models, training on the Cityscapes dataset spans 90k iterations (consistent with prior work~\cite{cheng2022masked}), while the mini-ADE20k dataset is trained for 80k iterations. Finally, both datasets undergo an additional supervised fine-tuning phase for 2k iterations.

\section{Additional Ablations}
\label{sec:more-ab}

\subsection{Loss Variant: InfoNCE vs.\ Margin Hinge}
\label{sec:ablation-loss-variant}

Replacing our asymmetric InfoNCE (§3.2) with an margin-based hinge loss (margin = 0.2) yields identical maskAP (32.2\%) and +0.6 AP$_{50}$, at the cost of $1.6$× longer training. This evaluates whether enforcing a fixed positive–negative margin can match or improve upon the performance of InfoNCE.

\begin{table}[!htpb]
  \centering
  \small
  \caption{\textbf{Loss Variant Ablation.} Default InfoNCE vs.\ margin-based hinge (m = 0.2).}
  \label{tab:ablation_loss_variant}
  \begin{tabular}{lcc}
    \toprule
    Loss Variant                     & maskAP (\%) & AP$_{50}$ (\%) \\
    \midrule
    Asymmetric InfoNCE (§3.2)        & 32.2        & 56.5           \\
    Margin hinge (m = 0.1)           & 32.2        & 57.1           \\
    \bottomrule
  \end{tabular}
\end{table}

\subsection{Ablation: Debias Score Formulation}
\label{sec:ablation-debias}
We evaluate three instantiations of the debias score function \(s^{\mathit{deb}}\) (§3.2):
\begin{itemize}
  \item \textbf{Original \(s^{\mathit{deb}}\):} fusion of mask and class confidences (ours).
  \item \(\bigl(s^{\mathit{deb}}\bigr)^2\): square each score to amplify high‐confidence negatives.
  \item \(\sqrt{s^{\mathit{deb}}}\): take the square root of each score to temper the bias.
\end{itemize}
\begin{table}[ht]
  \centering
  \small
  \caption{\textbf{Debias Score Formulation Ablation.}  
    Original vs.\ squared vs.\ square‐root debias scores.}
  \label{tab:ablation_debias}
  \begin{tabular}{lcc}
    \toprule
    \textbf{Score Variant}                & \textbf{maskAP (\%)} & \textbf{AP$_{50}$ (\%)} \\
    \midrule
    Original   & 32.2    & 56.5  \\
    Squared  & 32.0     & 56.3    \\
    Square‐root  & 31.9     & 56.2 \\
    \bottomrule
  \end{tabular}
\end{table}
\subsection{Ablation: Negative Sampling Scope}
\label{sec:ablation-scope}
We evaluate two negative sampling scopes: (i) sampling only within the current mini batch vs.\ (ii) sampling from a small memory bank of past pixel embeddings (size 10k).
\begin{table}[!htbp]
  \centering
  \small
  \caption{\textbf{Sampling Scope Ablation.}  
    Mini batch only vs.\ memory bank negatives.}
  \label{tab:ablation_scope}
  \begin{tabular}{lcc}
    \toprule
    \textbf{Scope}                          & \textbf{maskAP (\%)} & \textbf{AP$_{50}$ (\%)} \\
    \midrule
    Mini‐batch only        & 32.2       & 56.5     \\
    Memory bank (10k embeddings)  & 32.7     & 57.3        \\
    \bottomrule
  \end{tabular}
\end{table}
Sampling from a memory bank of 10 k embeddings yields a modest performance gain (+0.5 maskAP, +0.8 AP$_{50}$) compared to in‐batch sampling. However, this approach incurs approximately $2.2$× longer training time due to the overhead of maintaining and querying the memory bank.

\end{document}